\pdfoutput=1

\documentclass[11pt]{article}

\usepackage[]{ACL2023}

\usepackage{times}
\usepackage{latexsym}
\usepackage{color}
\usepackage{tabularray}
\usepackage{amsmath} 
\usepackage{colortbl}
\usepackage{amsmath,array,graphicx}
\usepackage{kantlipsum}
\usepackage{color}
\usepackage{adjustbox}
\usepackage{tabularray}

\usepackage[T1]{fontenc}

\usepackage[utf8]{inputenc}

\usepackage{microtype}

\usepackage{inconsolata}

\title{SQL-Encoder: Improving NL2SQL In-Context Learning Through a Context-Aware Encoder}

\author{
Mohammadreza Pourreza\textsuperscript{1}\textsuperscript{2}, 
Davood Rafiei\textsuperscript{2},
Yuxi Feng\textsuperscript{1}, 
Raymond Li\textsuperscript{1}, 
Zhenan Fan\textsuperscript{1}, 
Weiwei Zhang\textsuperscript{1} \\
\textsuperscript{1}Huawei Technologies Canada Co. Ltd, \textsuperscript{2}University of Alberta, \\
\texttt{\{mohammadreza.pourreza, yuxi.feng, raymond.li2, zhenan.fan1, weiwei.zhang2\}@huawei.com} \\
\texttt{drafiei@ualberta.ca}
}

\begin{document}
\maketitle
\begin{abstract}
Detecting structural similarity between queries is essential for selecting examples in in-context learning models. However, assessing structural similarity based solely on the natural language expressions of queries, without considering SQL queries, presents a significant challenge. This paper explores the significance of this similarity metric and proposes a model for accurately estimating it. To achieve this, we leverage a dataset comprising 170k question pairs, meticulously curated to train a similarity prediction model. Our comprehensive evaluation demonstrates that the proposed model adeptly captures the structural similarity between questions, as evidenced by improvements in Kendall-Tau distance and precision@k metrics. Notably, our model outperforms strong competitive embedding models from OpenAI and Cohere. Furthermore, compared to these competitive models, our proposed encoder enhances the downstream performance of NL2SQL models in 1-shot in-context learning scenarios by 1-2\% for GPT-3.5-turbo, 4-8\% for CodeLlama-7B, and 2-3\% for CodeLlama-13B.
\end{abstract}

\section{Introduction}

Scaling language models (LMs) has impressively enhanced sample efficiency and performance across numerous natural language processing (NLP) tasks \citep{brown2020language}. A notable area of improvement is in translating natural language questions into SQL (NL2SQL) queries. In this domain, models leveraging in-context learning, which involves providing a few example question and SQL pairs in the prompt, have achieved state-of-the-art performance \citep{pourreza2023din, gao2023text} on NL2SQL benchmarks \citep{yu2018spider, li2023llm}. In-context learning, a capability that has emerged as a feature of large language models (LLMs) \citep{wei2022emergent}, enables LLMs to adapt to specific tasks with a few examples. This paper focuses on enhancing the in-context learning performance of LLMs for the NL2SQL task. We propose a novel method for effectively measuring the similarity between task-specific demonstrations, alongside a model trained to predict this similarity score.

In the NL2SQL domain, the majority of prior research \citep{liu2021makes, nan2023enhancing, guo2023case, gao2023text} has concentrated on cross-domain in-context learning. This approach typically involves selecting few-shot examples from question and SQL pairs derived from databases different from the one associated with the question at inference time. We contend that relying on cross-domain example selection fails to unlock the full capabilities of in-context learning in NL2SQL domain. This is because examples from other databases may not effectively inform the generation of SQL queries for a given question, especially when the underlying tables and columns differ significantly. Table \ref{tab:1} showcases the one-shot performance of CodeLlama 13B \citep{roziere2023code} on the BIRD \cite{li2023llm} dataset development set under two scenarios: one utilizing cross-domain examples and the other employing in-domain examples, where the in-context learning samples are from the exact same database as the question at inference time. Selection was based on embedding all examples and the given question with the OpenAI embedding model, then using cosine similarity to identify the most relevant example. The findings indicate a significant 12\% enhancement in SQL generation performance when utilizing a single in-domain example, in contrast to cross-domain examples that appear to diminish the efficacy of one-shot in-context learning. This observation is in line with \citet{gao2023text}, which noted a reduction in performance of the GPT-3.5-turbo model under one-shot cross-domain conditions~\footnote{It should be noted that in cross-domain scenarios, prior research \citep{liu2021makes, nan2023enhancing, guo2023case, gao2023text} has demonstrated that employing a few-shot approach with more than one sample can ultimately enhance performance, outperforming the zero-shot scenario.}. This paper aims to explore \textit{in-domain in-context learning scenarios}, seeking to enhance performance in a way that truly reflects the strengths of in-context learning.

The prevalent strategy for in-context learning with LLMs involves deriving question embeddings through high-performing embedding models, such as the top embedding models on MTEB \citep{muennighoff2022mteb} benchmark. This technique entails measuring cosine similarity or dot product among the embeddings to identify the most closely related questions \citep{guo2023case, liu2021makes}. However, recent studies \citep{nan2023enhancing, gao2023text} have shown that relying solely on question similarities may not always lead to the selection of the highest performing in-context learning samples, in terms of the accuracy of the generated SQL queries. Instead, these investigations have considered SQL query skeletons as an additional layer of information for finding samples for in-context learning. A key question investigated in this paper is if the structure of SQL queries and their similarity to other queries can be accurately predicted based solely on questions, without the need to generate SQL queries.
This paper explores the significance of such similarity metrics and proposes novel measures, such as schema-linking similarity and SQL skeleton similarity with tree edit distance, to enhance in-context learning. Upon identifying the most effective similarity metric, we created a large dataset of 170K question pairs and their respective similarity scores. This dataset was used to train a cross-encoder language model \citep{reimers2019sentence}, specifically designed to predict the similarity score between a pair of questions \footnote{ model is accessible in huggingface: https://huggingface.co/MrezaPRZ/sql-encoder}. This cross-encoder model, the first of its kind tailored for the NL2SQL task, has shown in our experiments to provide superior signals for selecting few-shot samples, outperforming both OpenAI \citep{neelakantan2022text, openai} and Cohere's embedding \citep{cohere} models for in-context learning task. Its exceptional performance, further validated on out-of-domain questions, surpasses that of the OpenAI and Cohere models on such samples, demonstrating our model's generalizability.

\definecolor{Gray}{rgb}{0.501,0.501,0.501}
\definecolor{MineShaft}{rgb}{0.2,0.2,0.2}
\begin{table}
\centering
\begin{tblr}{
  cells = {c},
  row{1} = {Gray},
  cell{2}{1} = {fg=MineShaft},
}
\textbf{Method} & \textbf{Execution accuracy}\\
Zero-shot & 26.66\\
In-domain one-shot & 38.14\\
Cross-domain one-shot & 20.99
\end{tblr}
\caption{Performance comparison of the in-domain and cross-domain in-context learning on the development set of BIRD}
\label{tab:1}
\end{table}

\begin{figure*}[ht]
\centering
  \includegraphics[width=0.85\textwidth]{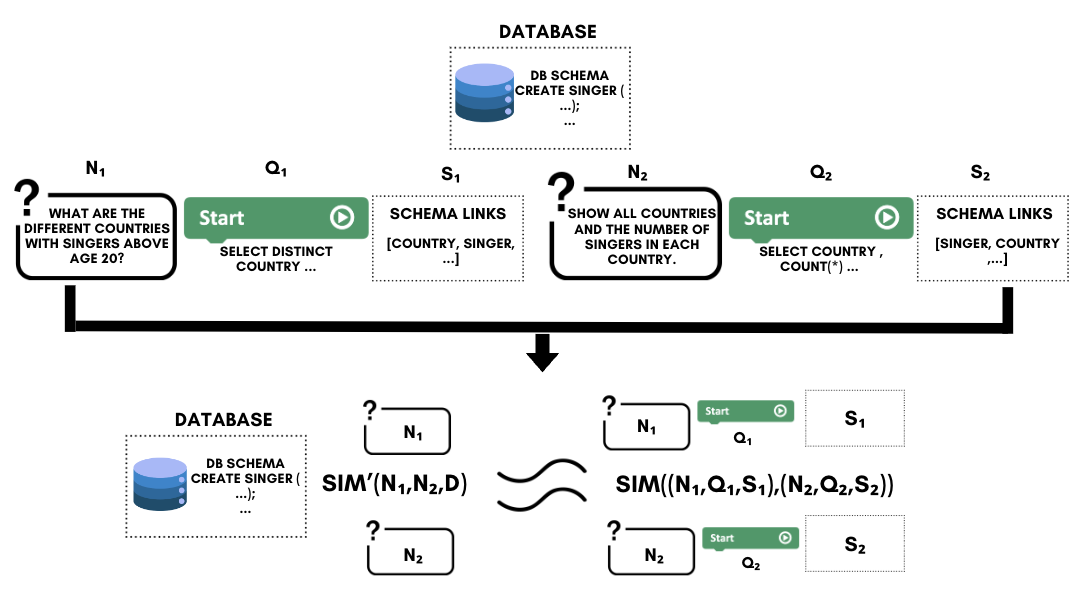}
  \caption{Overview of SQL-encoder framework for predicting the similarity between two questions $N_1$ and $N_2$ on a given database schema D, where SIM' serves as a proxy for the actual similarity SIM between the questions $N_1$ and $N_2$, their respective SQL queries $Q_1$ and $Q_2$ and schema links $S_1$ and $S_2$.}
  \label{fig:1}
\end{figure*}

\section{Methodology}

In this paper, our primary objective is to identify the most effective similarity metric to enhance the performance of in-context learning. Specifically, we aim to estimate the similarity between two queries $Q_1$ and $Q_2$ given their natural language expressions $N_1$ and $N_2$ and a Schema $S$. Our initial step involves identifying the function that yields the highest performance in in-context learning. However, during testing, instead of having access to $Q_1$ and $Q_2$, we work with natural language questions $N_1$ and $N_2$ alongside Schema $S$. Here, $Q_1$ and $Q_2$ represent SQL queries on Schema S that respectively answer $N_1$ and $N_2$. We have constructed a dataset comprising pairs of queries and have labelled the proxy function sim($N_1$,$N_2$, $S$) with the actual similarity sim(($N_1$,$Q_1$,$S_1$),($N_2$,$Q_2$,$S_2$)) that we want to approximate, where $S_1$ and $S_2$ are schema links of $N_1$ and $N_2$. This approach allows us to train a model to predict this proxy similarity measure. Figure \ref{fig:1} outlines the methodology adopted in this study to train the model using the optimal similarity metric.

\subsection{NL2SQL Similarity Metric}

\begin{figure*}[ht]
\centering
  \includegraphics[width=0.8\textwidth]{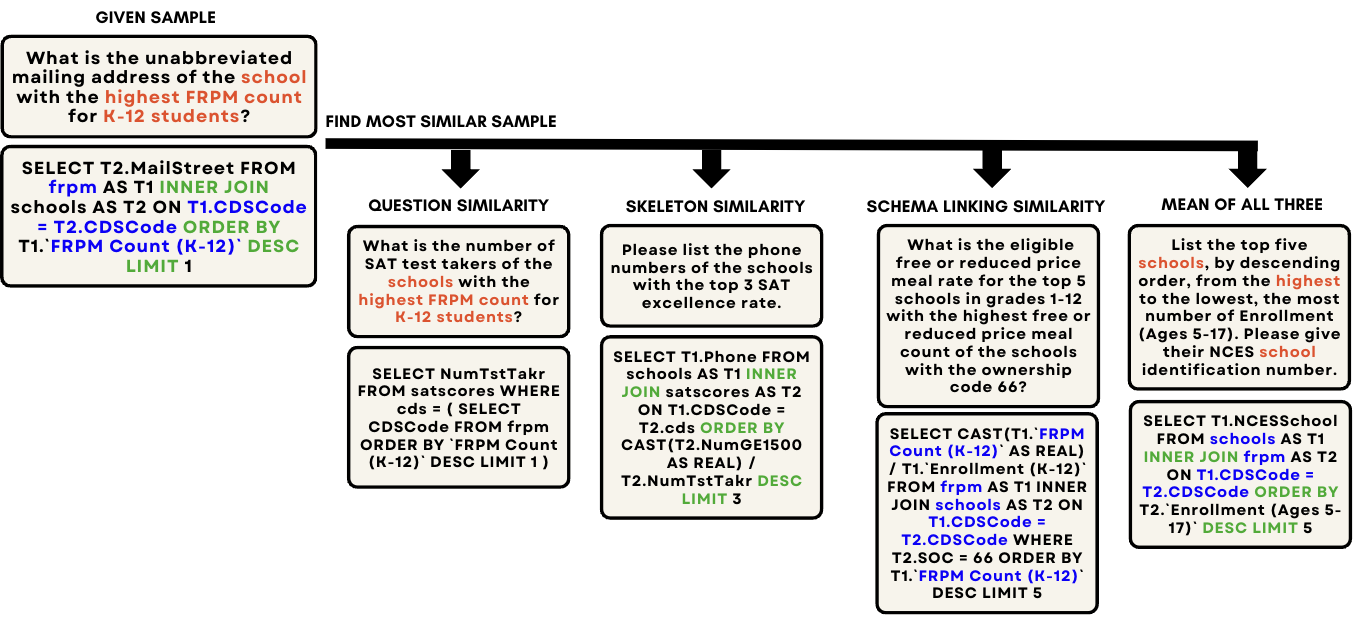}
  \caption{An example demonstrating the utilization of different similarity metrics to find the most similar Question/SQL pair.}
  \label{fig:2}
\end{figure*}

In previous studies \citep{nan2023enhancing, gao2023text}, it has been shown that relying exclusively on question similarity to select in-context samples leads to sub-optimal performance. In our research, we aim to compare various NL2SQL-specific similarity metrics and elucidate their contributions towards enhancing in-context learning performance. Figure \ref{fig:2} demonstrates an example of using different similarity metrics to find the most similar Question/SQL pair for a given Question/SQL pair from the development set of BIRD dataset. 

\subsubsection{Question Similarity}

Similar questions often reference the same table and column names, or employ identical keywords when ordering or aggregating results. Therefore, the most straightforward approach to identifying in-context learning samples is by leveraging question similarity.
This method may embed all questions within the sample pool, and then, for any specific question, after its embedding, a similarity metric such as cosine similarity is applied to identify samples with the most similar question embeddings. However, relying solely on question similarity is inadequate because NL2SQL requires structured prediction, which necessitates more explicit information about the problem's structure than what is provided by the input question alone \citep{nan2023enhancing}.

\subsubsection{Query Skeleton Similarity}

Recent studies by \citet{nan2023enhancing, gao2023text} introduced the concept of utilizing SQL query skeleton similarity as a criterion for selecting in-context learning samples. This method aims to guide the model about the necessary syntax and structure for crafting SQL queries. Their methodology starts with generating a draft SQL query for the given question using a preliminary NL2SQL model, then, encoding both the draft query and all queries from the sample pool into binary syntax vectors according to SQL keyword presence, and then selecting the most similar queries as in-context learning samples. While promising, this approach, we argue, falls short by neglecting the structural and keyword order intricacies inherent to SQL queries, resembling the limitations found in the bag-of-words model \citep{hiz2012papers}. Moreover, when the preliminary NL2SQL model produce a draft SQL that significantly diverges from a correct solution, this method fails to accurately assess similarity.

\begin{figure}[h]
\centering
  \includegraphics[width=0.5\textwidth]{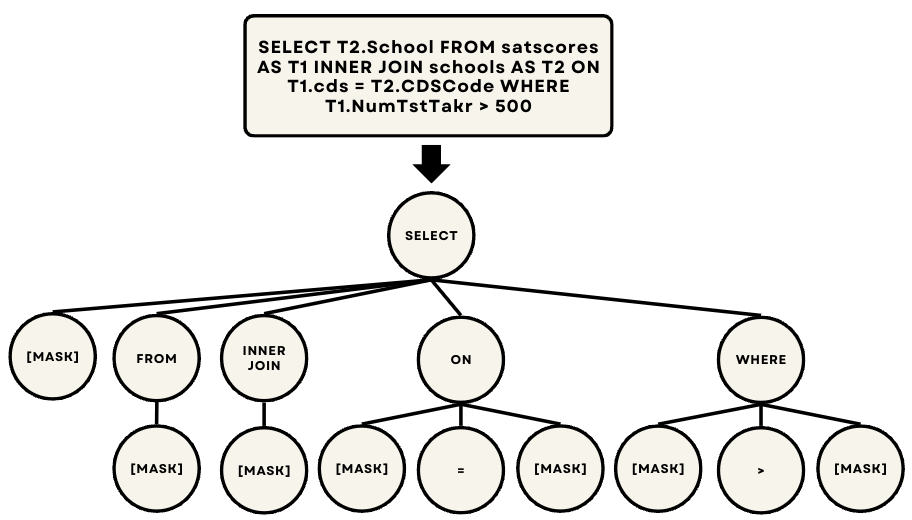}
  \caption{An example of the process to construct a tree from SQL query after masking the schema mentions.}
  \label{fig:3}
\end{figure}

To address these shortcomings, our work proposes the construction of abstract syntax trees from SQL query skeletons, as shown in Figure \ref{fig:3}, employing tree edit distance \citep{zhang1989simple} to measure the query dissimilarity. In order to turn this dissimilarity metric into a similarity metric between 0 and 1, we normalized the distance by dividing the distances by the maximum distance for all of the pairs and using 1 - distance as the similarity metric. This method maintains the structural integrity of SQL queries and surpasses the performance of the syntax vector approach, showcasing its efficacy in preserving the order and structure of SQL query keywords.

\subsubsection{Schema-Linking Similarity}

Previous research in the NL2SQL field \citep{pourreza2023din, li2023llm, pourreza2024dts, dominguez2024blar} has highlighted that a significant number of failures in NL2SQL models stem from mismatches between the schema links (tables and columns) in generated SQL queries and those in reference SQL queries. Contrary to cross-domain in-context learning, in-domain settings allow the utilization of table and column names as additional criteria for selecting in-context samples. In our study, we introduce a novel approach tailored to in-domain in-context learning, which selects samples based on the Jaccard similarity of the sets of tables and columns used within the queries.

\subsubsection{Mean Of All Three}

Each of the aforementioned similarity metrics captures a unique aspect of similarity between two question/SQL pairs. To holistically assess the similarity score across all these dimensions, we calculate the mean of all three metrics as a definitive measure for selecting in-context samples. This integrated approach yielded superior in-context learning performance in our experiments. From now on, when we mention a similarity score, we're referring to the combined score from all three metrics.

\subsection{Encoder}
Following an extensive evaluation of various similarity metrics to identify the most efficacious approach, we utilized the BIRD dataset \citep{li2023llm} to compile a comprehensive finetuning dataset. This dataset comprises 170,000 pairs of questions, each annotated with a corresponding similarity score, serving as our supervised finetuning corpus. To accommodate the requisite knowledge of the database schema, we constructed in the input sequence to the model as the concatenation of the database schema together with the first and second question. Given that our task is a regression task, where the objective is to predict a continuous number between 0 and 1, we have applied a sigmoid function to the logits output by the model. This ensures that the predictions are confined within the 0 to 1 range. Subsequently, we utilized the Mean Absolute Error (MAE) as the loss function:

\begin{equation}
\text{L} = \frac{1}{N} \sum_{i=1}^{N} \left| y_i - \frac{1}{1 + e^{-z_i}} \right|
\end{equation}

\noindent where $N$ is the number of samples, $y_i$ denotes the actual similarity score for the $i$-th question pair, and $z_i$ represents the predicted similarity by the model for the $i$-th question pair.

Contrary to previous methodologies that employed cross-encoder or ranking models built on pretrained architectures such as BERT \citep{devlin2018bert} or ROBERTA \citep{liu2019roberta}, investigated by \citep{nogueira2019multi, reimers2019sentence, chen2024bge}, our approach necessitated the handling of larger input sequences exceeding 2096 tokens. This requirement stemmed from integrating a database schema with two distinct questions. Additionally, our procedure could be enhanced by a code pretraining phase, akin to the pretraining seen in models like CodeLlama \citep{roziere2023code} or DeepSeek \citep{guo2024deepseek}. Nevertheless, given the need for cross-encoder models to compute similarity scores between each pair efficiently, the model's size and speed are critical considerations. Therefore, we opted for the smallest version of the DeepSeek models, which has 1.3 billion parameters, as our foundational model for similarity score prediction. In order to make a single number prediction, a single linear layer with an input feature size of 2048 and an output size of 1 has been attached to the output of the last token from the model.

\section{Experiments}

\subsection{Models}

To evaluate the effectiveness of our trained encoder model in predicting similarity scores for in-domain in-context learning sample selection, we report the performance of the in-context learning on three models from distinct families and architectures serving as SQL generators. Specifically, we utilized the CodeLLama 7B and 13B models \citep{roziere2023code} alongside the gpt-3.5-turbo model, to assess the performance of the in-context learning.

\subsection{Datasets}
Our evaluation involved two comprehensive cross-domain datasets, Spider and BIRD \citep{yu2018spider, li2023llm}. Spider contains 10,181 questions linked to 5,693 SQL queries across 138 domains and 200 databases, divided into 8,659 training, 1,034 development, and 2,147 test examples from unique databases. BIRD comprises 12,751 question-SQL pairs from 95 databases across 37+ domains, including blockchain and healthcare, totaling 33.4 GB. The dataset breakdown for BIRD includes 1,534 development, 9,428 training, and 1,789 test queries. We used the development sets from both datasets for evaluation without performing hyperparameter tuning.

To create the training dataset, we utilized BIRD's training set. For each question, we sampled 20 (or the total number available questions per database if it was fewer than 20) other questions from the same database. We then calculated the mean of all three similarity metrics for the pair to serve as the prediction label. This approach resulted in a training dataset of roughly 170K question pairs, which was used to train our model.

\subsection{Metrics}
In the NL2SQL field, model performance is mainly assessed using two metrics: exact set match accuracy and execution accuracy. Exact set match accuracy evaluates the alignment of components like SELECT, WHERE, and GROUP BY in both generated and reference SQL queries without considering their sequence. Execution accuracy, on the other hand, measures whether a query produced by the model and the reference query return the same results across various databases. Given the diversity in correct SQL query formulations for a single question, execution accuracy is considered a more robust metric. This metric has been used as the primary metric for both Spider and BIRD benchmarks. Our study also adopts execution accuracy as the primary evaluation metric.

\subsection{Hyperparameters}

We fine-tuned the DeepSeek model with 1.3B parameter on 160K samples for three epochs, setting aside 10K samples for validation and finding the best checkpoint. A full parameter finetuning was executed at a learning rate of 5e-5, using the AdamW optimizer \citep{loshchilov2017decoupled} and a batch size of 8, on a single H100 GPU.

\subsection{Results}

\subsubsection{Encoder Similarity Prediction}

This section aims to assess the encoder model's performance in predicting similarity scores. These scores are evaluated in terms of being aligned with the mean similarity scores obtained by averaging three similarity metrics, using the ground truth SQL query.

To evaluate the similarity prediction score, we employed the Kendall Tau coefficient \citep{kendall1938new} to compare the orderings generated by our model's predicted scores against the orderings derived from the mean of the three similarity metrics. The Kendall Tau coefficient ranges between -1 and 1, where 1 indicates a perfect match between the orderings, and -1 indicates a completely opposite ordering. Additionally, given the significance of the highest-scoring samples in few-shot in-context learning—since these samples are selected for inclusion in the prompt, we also calculated the precision in identifying the top 1, 5, 10, 15, and 20 samples at the top 1, 5, 10, 15, and 20 results predicted by our encoder.

Table \ref{tab:7} presents the mean Kendall Tau coefficients and precision@k for different values of $k$ on the Spider dataset's development set, comparing the orderings from the best similarity metric (mean of three) with those obtained by our encoder, the orderings based on question similarity (using the OpenAI embedding model), the orderings based on SQL skeleton similarity, and the orderings based on schema-linking similarity together with the precision@10. It is important to note that for the last two similarity metrics, we utilize the ground truth SQL query for similarity prediction, hence they serve as upper bounds. However, our trained encoder model was designed to predict the mean of three similarities without access to the SQL query, relying solely on the questions.

\begin{table*}
\centering
\begin{tblr}{
  row{even} = {c},
  row{1} = {Gray,c},
  row{3} = {c},
  cell{5}{2} = {c},
  cell{5}{3} = {c},
  cell{5}{4} = {c},
  cell{5}{5} = {c},
  cell{5}{6} = {c},
  cell{5}{7} = {c},
}
\textbf{Method} & \textbf{Mean KT} & \textbf{Mean P@1} & \textbf{\textbf{Mean P@5}} & \textbf{\textbf{Mean P@10}} & \textbf{\textbf{Mean P@15}} & \textbf{\textbf{Mean P@20}}\\
Encoder & \textbf{0.126} & \textbf{0.52} & 0.62 & \textbf{0.69} & \textbf{0.73} & \textbf{0.76}\\
Schema-linking & 0.115 & 0.26 & \textbf{0.63} & \textbf{0.69} & 0.71 & 0.72\\
SQL Skeleton & 0.060 & 0.16 & 0.38 & 0.46 & 0.54 & 0.6\\
Question-similarity & 0.100 & 0.41 & 0.58 & 0.65 & 0.69 & 0.73
\end{tblr}
\caption{Mean Kendall Tau coefficients and precision@k for different values of k,  comparing the ordering of different similarity metrics to that of the mean of all three metric on the development set of Spider.}
\label{tab:7}
\end{table*}

The findings presented in Table \ref{tab:7} reveal that our encoder aligns more closely with the ordering derived from the mean of the three similarity metrics than any individual metric that contributes to the mean. Furthermore, in terms of precision, our encoder is generally outperforming or reaching the same performance as all of the other three metrics. We believe that both precision@k and the Kendall Tau coefficient could be enhanced by employing larger base models. This is because predicting similarity by inferring the resemblance between SQL queries is a complex task, and the model should infer the SQL query internally in order to predict the similarity score and even the largest NL2SQL models like GPT-4 still struggle to perfectly predict the SQL query  \citep{li2023llm}. However, a trade-off exists, as larger models tend to slow down the similarity prediction process. This trade-off is precisely why we opted for a smaller base model.

\subsubsection{Down-Stream Performance of Different Similarity Metrics}

In this section, we evaluate different methods for measuring the similarity between in-context Question/SQL pairs, with the goal of finding the metric that achieves the best down-stream performance. This evaluation is done on the development set of BIRD. Having a pool of Question/SQL pairs from a database and a given question from the same database at inference time, in order to find the best similarity metric, we assume that we have access to the SQL query to measure the schema-linking and SQL skeleton similarity. This assumption was made to find an upper bound for the similarity metrics. Previous works \citet{nan2023enhancing, gao2023text} used a preliminary NL2SQL model to predict a draft SQL query and used that as the SQL query to measure the SQL skeleton similarity. After finding the best metric, our trained model only relies on the input questions and the database schema and is supposed to infer the SQL query skeleton from the input questions and database schema. The results of the comparison between different similarity metrics are reported in Table \ref{tab:2} by using the gpt-3.5-turbo model as the SQL prediction model in a one-shot setting. For the question similarity method, we used OpenAI embedding model \citep{openai} to obtain the question embeddings.

\definecolor{Gray}{rgb}{0.501,0.501,0.501}
\definecolor{MineShaft}{rgb}{0.2,0.2,0.2}
\begin{table}
\centering
\begin{tblr}{
  cells = {c},
  row{1} = {Gray},
  cell{2}{1} = {fg=MineShaft},
}
\textbf{Method} & \textbf{EX}\\
Zero-shot & 42.96\\
Question similarity (1-shot) & 48.89\\
SQL-Skeleton syntax vector (1-shot) & 46.28\\
SQL-Skeleton tree edit distance (1-shot) & 48.57\\
Schema-linking similarity (1-shot) & 50.59\\
Mean of all three (1-shot) & 52.93
\end{tblr}
\caption{Performance of different similarity metrics in selecting  one-shot in-context learning examples for NL2SQL with gpt-3.5-turbo-1106 used as the LLM to generate SQL queries on the development set of BIRD.}
\label{tab:2}
\end{table}

The data in Table \ref{tab:2} underscores the superiority of our proposed approach, which averages question similarity, schema-linking similarity, and SQL query skeleton similarity using the tree edit distance method. The proposed similarity metric can outperforms the best single similarity metric by 2.34\%. Furthermore, the tree edit distance showcases enhanced performance over the basic syntax vector similarity method mentioned in \citet{nan2023enhancing}, emphasizing the significance of acknowledging the structure and order of SQL keywords in the queries.

\subsubsection{Encoder Performance on BIRD}

\label{bird:results}

In this section, we compare the performance of our encoder model, which utilizes the database schema and two questions as inputs, with the question similarity approaches that rely on leading embedding models such as OpenAI and Cohere \citep{openai, cohere}. It is important to note that we are not comparing our encoder's performance with the methods outlined in \citet{nan2023enhancing, gao2023text}, which assess SQL Skeleton similarity by employing a preliminary model to predict a draft SQL query. These studies have demonstrated that the best performance is achieved when question similarity is integrated with SQL skeleton similarity. The decision to exclude their performance from this comparison is twofold: firstly, our trained model is primarily a question similarity method, as it processes questions directly rather than draft SQL queries; and secondly, our method is orthogonal to these works and our trained model could be used in their proposed framework to predict the question similarity. Table \ref{tab:3} showcases the performance of our encoder model in comparison to other question similarity approaches that utilize OpenAI text-embedding-3-large and Cohere cohere-embed-english-v3 embedding models \citep{openai, cohere}.

\definecolor{Gray}{rgb}{0.501,0.501,0.501}
\definecolor{MineShaft}{rgb}{0.2,0.2,0.2}
\begin{table}
\centering
\begin{tblr}{
  cells = {c},
  row{1} = {Gray},
  cell{2}{1} = {fg=MineShaft},
  cell{5}{1} = {fg=MineShaft},
  cell{8}{1} = {fg=MineShaft},
}
\textbf{Encoder} & \textbf{Model} & \textbf{EX}\\
Encoder & gpt-3.5-turbo & 50.46\\
OpenAI & gpt-3.5-turbo & 48.89\\
Cohere & gpt-3.5-turbo & 49.93\\
Zero-shot & gpt-3.5-turbo & 42.96 \\
Encoder & CodeLlama-7B & 38.59\\
OpenAI & CodeLlama-7B & 34.35\\
Cohere & CodeLlama-7B & 33.18\\
Zero-shot & CodeLlama-7B & 24.77\\
Encoder & CodeLlama-13B & 40.09\\
OpenAI & CodeLlama-13B & 38.14\\
Cohere & CodeLlama-13B & 37.16 \\
Zero-shot & CodeLlama-13B & 26.66
\end{tblr}
\caption{Performance of different question-similarity prediction models in 1-shot in-conext learning for NL2SQL task. For Cohere and OpenAI embedding models an extra step of cosine similarity calculation is required.}
\label{tab:3}
\end{table}

The results displayed in Table \ref{tab:3} highlight the superior performance of our question-similarity prediction model compared to conventional models designed for assessing the similarity between two questions. Unlike the other two embedding models, where the similarity score is derived using cosine similarity between the embedding vectors, our encoder directly predicts the similarity score. Our trained encoder has consistently outperformed the other two models across various models for SQL prediction in one-shot setting, demonstrating its effectiveness.

\subsubsection{Encoder Performance Across BIRD Query Classes}

\label{bird:complexity}

The BIRD dataset categorizes question/SQL pairs into three distinct classes: simple, moderate, and challenging, based on the complexity of the target SQL queries \citep{li2023llm}. 
In this section, we compare the performance across different query difficulty levels to explore the performance gains for each class. Table \ref{tab:4} demonstrates the performance across classes using different question similarity methods on various models, including gpt-3.5-turbo-1106, codellama-7b, and codellama-13b for in-context learning on the development set of the BIRD dataset.

\begin{table}
\centering
\begin{tblr}{
  cells = {c},
  row{1} = {Gray},
}
\textbf{Encoder} & \textbf{Model} & \textbf{Sim} & \textbf{Mod} & \textbf{Cha}\\
Encoder & GPT & 59.26 & 39.78 & 28.47\\
OpenAI & GPT & 59.25 & 33.55 & 31.25\\
Cohere & GPT & 58.38 & 39.35 & 29.86\\
Encoder & CL-7B & 47.78 & 25.38 & 22.22\\
OpenAI & CL-7B & 44 & 21.29 & 14.58\\
Cohere & CL-7B & 41.41 & 22.58 & 14.58\\
Encoder & CL-13B & 49.62 & 28.17 & 17.36\\
OpenAI & CL13B & 46.92 & 26.88 & 18.06\\
Cohere & CL-13B & 45.08 & 27.31 & 18.06
\end{tblr}
\caption{Performance  of various question-similarity prediction models in one-shot in-context learning for the NL2SQL task across simple (Sim), moderate (Mod), and challenging (Cha) difficulty levels on the BIRD development set. 'GPT' refers to gpt-3.5-turbo, and 'CL' denotes CodeLlama.}
\label{tab:4}
\end{table}

Based on the results, it is evident that our encoder model consistently surpasses the in-context learning performance of two other models for question similarity, particularly for smaller models such as CodeLlama 7B, with the most significant performance improvement observed in challenging samples. In contrast, with larger models like gpt-3.5-turbo and CodeLlama 13B, our encoder model excels in the simple and moderate categories but does not perform as well as the two other embedding models in the challenging category. We believe that the performance enhancement of our encoder model is particularly noticeable in scenarios where in-context learning yields more substantial improvements over zero-shot performance, as can also be seen in Table \ref{tab:3}.

\subsubsection{Encoder Performance on Spider}

Given that our model was trained on the BIRD training set, concerns regarding the generalizability of our question-similarity prediction model may arise. To address these concerns, we evaluated our encoder model's performance on the Spider development set, comparing it with other question-similarity prediction methods that utilize OpenAI's text-embedding-3-large and Cohere's cohere-embed-english-v3 embedding models \citep{openai, cohere}. For SQL prediction, we employed the CodeLlama 13B parameter model to assess its one-shot performance. The results are presented in Table \ref{tab:4}, highlighting the superior performance of our encoder model with almost 4\% gain for both exact set match and execution accuracy over the other methods. One interpretation of the observed improvement in exact set match over the other two methods could be that our encoder preferentially selects samples with SQL queries more similar to the target SQL query.

\definecolor{Gray}{rgb}{0.501,0.501,0.501}
\definecolor{MineShaft}{rgb}{0.2,0.2,0.2}
\begin{table}
\centering
\begin{tblr}{
  cells = {c},
  row{1} = {Gray},
  cell{2}{1} = {fg=MineShaft},
}
\textbf{Encoder} & \textbf{Model} & \textbf{EX} & \textbf{EM} \\
Encoder & CodeLlama-13B & 73.2 & 50\\
OpenAI & CodeLlama-13B & 69.1 & 46.3\\
Cohere & CodeLlama-13B & 69.3 & 46.3\\
Zero-shot & CodeLlama-13B & 64.7 & 24.9
\end{tblr}
\caption{Performance of the in-domain in-context learning on Spider development set using question similarity. EX refers to execution accuracy and EM refers to exact set match accuracy.}
\label{tab:5}
\end{table}

\subsubsection{Encoder Performance Across Spider Query Classes}

The Spider dataset categorizes its Question/SQL pairs into four complexity classes: easy, medium, hard, and extra hard, based on the complexity of the structured SQL queries. Mirroring our analysis in Section \ref{bird:complexity}, we compare our question similarity predictor encoder model against two other embedding models across these complexity classes on the Spider development set, utilizing CodeLlama-13B as the SQL prediction model. The results are detailed in Table \ref{tab:6}.

\begin{table}
\centering
\begin{tblr}{
  cells = {c},
  row{1} = {Gray},
}
\textbf{Encoder} & \textbf{Model} & \textbf{Easy} & \textbf{Med} & \textbf{Hard} & \textbf{Ext}\\
Encoder & CL-13B & 88.3 & 80.9 & 55.7 & 48.2\\
OpenAI & CL13B & 84.7 & 75.3 & 55.2 & 44.0\\
Cohere & CL-13B & 85.5 & 75.8 & 55.2 & 42.8
\end{tblr}
\caption{Performance  of various question-similarity prediction models in one-shot in-context learning for the NL2SQL task across easy, medium, hard, and extra hard difficulty levels on the Spider development set. 'CL' denotes CodeLlama.}
\label{tab:6}
\end{table}

Our encoder model consistently outperforms the other two embedding models in question similarity prediction across all complexity levels, achieving the most significant gains in the extra hard and medium classes.

\section{Conclusion}
Embedding questions into vectors transcends term-based methods, enabling the detection of questions that may be similar but expressed differently. However, generic embedding vectors cannot capture all intricacies of questions and how they may be formulated in SQL. Our work is the first to develop a similarity function for identifying structural similarities between queries based on their natural language expressions. Our approach, utilizing a lightweight model with 1.3 billion parameters, surpasses the performance of the high performing embedding models across both Bird and Spider datasets. The pursuit of specialized functions to better discern structural similarities between queries and their integration into NL2SQL pipelines shows promise for enhancing model performance.

\section*{Limitations}

Contrary to embedding models that yield a single embedding vector suitable for storage and subsequent reuse for computing similarities between question pairs—thereby only necessitating the embedding of new samples for distance calculations against pre-existing vectors—our methodology adopts a cross-encoder architecture. This architecture requires the processing of each question pair through the similarity prediction model, rendering it less efficient for contexts characterized by extensive sets of question pairs. This design decision was made due two primary reasons. Firstly, the accurate prediction of SQL query similarities from questions demands an analytical process that encompasses both inputs, a requirement effectively met by the cross-encoder architecture. Secondly, in-context learning is predominantly used for smaller sample sizes, finetuning methods are generally more advantageous for larger datasets. \citep{li2024codes,pourreza2024dts}

\section*{Ethics Statement}

In this paper, we emphasize the significance of ethical considerations in our research, ensuring compliance with the ACL Ethics Policy and adherence to ethical guidelines throughout. Efforts were made to minimize biases and discrimination in our methodology and analysis. We have committed to transparency, accuracy, and fairness in our findings, providing due citations and acknowledgments. This ethics statement underscores our dedication to integrity, respect for ethical standards, and our contribution to the responsible advancement of knowledge in our field.

\section*{Acknowledgements}

\bibliography{anthology,custom}
\bibliographystyle{acl_natbib}




\end{document}